# ROOT13: Spotting Hypernyms, Co-Hyponyms and Randoms

**Enrico Santus*, Alessandro Lenci§, Tin-Shing Chiu*, Qin Lu*, Chu-Ren Huang***

\* The Hong Kong Polytechnic University, Hong Kong
esantus@gmail.com, cstschiu@comp.polyu.edu.hk, {qin.lu, churen.huang}@polyu.edu.hk
§ University of Pisa, Italy
alessandro.lenci@ling.unipi.it

### Abstract

In this paper, we describe ROOT13, a supervised system for the classification of hypernyms, co-hyponyms and random words. The system relies on a Random Forest algorithm and 13 unsupervised corpus-based features. We evaluate it with a 10-fold cross validation on 9,600 pairs, equally distributed among the three classes and involving several Parts-Of-Speech (i.e. adjectives, nouns and verbs). When all the classes are present, ROOT13 achieves an F1 score of 88.3%, against a baseline of 57.6% (*vector cosine*). When the classification is binary, ROOT13 achieves the following results: hypernyms-co-hyponyms (93.4% vs. 60.2%), hypernyms-random (92.3% vs. 65.5%) and co-hyponyms-random (97.3% vs. 81.5%). Our results are competitive with state-of-the-art models.

## Introduction and Related Work

Distinguishing hypernyms (e.g. *dog-animal*) from co-hyponyms (e.g. *dog-cat*) and, in turn, discriminating them from random words (e.g. *dog-fruit*) is a fundamental task in Natural Language Processing (NLP). Hypernymy in fact represents a key organization principle of semantic memory (Murphy, 2002), the backbone of taxonomies and ontologies, and one of the crucial inferences supporting lexical entailment (Geffet and Dagan, 2005). Co-hyponymy (or coordination), on the other hand, is the relation held by words sharing a close hypernym, which are therefore attributionally similar (Weeds et al., 2014).

The ability of discriminating hypernymy, co-hyponymy and random words has potentially infinite applications, including automatic thesauri creation, paraphrasing, textual entailment, sentiment analysis and so on (Weeds et al., 2014). For this reason, in the last decades, numerous methods, datasets and shared tasks have been proposed to improve computers' ability in such discrimination, generally achieving promising results (Weeds et al., 2014; Rimmel, 2014; Geffet and Dagan, 2005). Both supervised and unsupervised approaches have been investigated. The former have been shown to outperform the latter in Weeds et al. (2014), even though Levy et al. (2015) have recently claimed that these methods may learn whether a term $y$ is a prototypical hypernym, regardless of its actual relation with a term $x$.

In this paper, we propose a supervised method, based on a Random Forest algorithm and 13 corpus-based features. In our evaluation, carried out using the 10-fold cross validation on 9,600 pairs, we achieved an accuracy of 88.3% when the three classes are present, and of 92.3% and 97.3% when only two classes are present. Such results are competitive with the state-of-the-art (Weeds et al., 2014).

## Method and Evaluation

ROOT13 uses the Random Forest algorithm implemented in Weka (Breiman, 2001), with the default settings. It relies on 13 features that are described below. Each of them is automatically extracted from a window-based Vector Space Model (VSM), built on a combination of ukWaC and WaCkypedia corpora (around 2.7 billion words) and recording word co-occurrences within the 5 nearest content words to the left and right of each target.

**FEATURES.** The feature set was designed to identify several distributional properties characterizing the terms in the pairs. On top of the standard features (e.g. vector cosine, co-occurrence and frequencies), we have added several features capturing the generality of the terms and of their contexts[1], plus two unsupervised measures for capturing similarity (Santus et al., 2014b-c). All the features are normalized in the range 0-1:

- **Cos**: vector cosine (Turney and Pantel, 2010);
- **Cooc**: co-occurrence frequency;
- **Freq 1, 2**: two features storing the frequency the terms;
- **Entr 1, 2**: two features storing the entropy of the terms;

---

[1] Generality is measured as for Santus et al. (2014a).

- **Shared**: extent of the intersection between the top 1k most mutually related contexts of the two terms[2];
- **APSyn**: for every context in the intersection between the top 1k most mutually related contexts of the two terms, this measure adds 1, divided by its average rank (Santus et al. 2014b-c);
- **Diff Freqs**: difference between the terms frequencies;
- **Diff Entrs**: difference between the terms entropies[3];
- **C-Freq 1, 2**: two features storing the average frequency among the top 1k most mutually related contexts for each term;
- **C-Entr 1, 2**: two features, storing the average entropy among the top 1k most mutually related contexts for each term.

**DATASET.** We have used 9,600 pairs, randomly extracted from three datasets: Lenci/Benotto (Santus et al., 2014b), BLESS (Baroni and Lenci, 2011) and EVALution (Santus et al., 2015). The pairs are equally distributed among the three classes (i.e. hypernyms, co-hyponyms and random words) and involve several Parts-Of-Speech.

**TASKS.** Four classification tasks have been carried out. One involving all classes and three tasks involving only two of them. F1 score on a 10-fold cross validation was chosen as accuracy measure.

**BASELINE.** Vector cosine is used as baseline. It achieves a reasonable accuracy, which is anyway far below the results obtained by our model. As it will be shown below, our model does not benefit from its use.

**RESULTS.** Table 1 describes the features' contributions in the four classification tasks. As can be seen, when all the classes are involved, every feature contributes for an increment between 0.4% and 4.8%, except for the feature **Shared**, whose contribution is +12.5%. The relevance of this feature is confirmed also in the other three tasks. Interestingly, instead, the vector cosine does not contribute to our score. It instead penalizes it in three tasks out of four. The only task in which it actually contributes for 0.1% is the discrimination between co-hyponyms and randoms, which is its main function.

## Conclusions

In this paper, we have described ROOT13, a classifier for hypernyms, co-hyponyms and random words. The classifier, based on the Random Forest algorithm, uses only 13 unsupervised corpus-based features, which have been described and their contribution reported. Our classifier is competitive with the state-of-the-art (Weeds et al., 2014). In a second run of tests, we have noticed the Levy et al. (2015)'s effect, that is the classification of switched hypernyms as hypernyms (e.g. *dog-vehicle*, *car-animal*). However, we were able to remove it – without any sensible loss in accuracy – by training the model also on switched hypernyms labeled as randoms.[4]

|  | Hyper Coord Random | Hyper Coord | Hyper Random | Coord Random |
|---|---|---|---|---|
| **Cos (Baseline)** | 57.6 | 60.2 | 65.5 | 81.5 |
| + Cooc | 62.4 | 71.9 | 71.7 | 81.4 |
| + Shared | 74.9 | 83.2 | 80.2 | 96.6 |
| + Diff Freqs | 76.6 | 83.6 | 82.6 | 96.2 |
| + Diff Entrs | 78.4 | 84.4 | 83.8 | 96.2 |
| + APSyn | 79.0 | 84.9 | 84.5 | 96.5 |
| + Freq 1, 2 | 83.1 | 88.2 | 88.7 | 96.6 |
| + Entr 1, 2 | 85.4 | 90.6 | 90.1 | 97.1 |
| + C-Entr 1, 2 | 87.6 | 92.6 | 92.1 | 97.3 |
| + C-Freq 1, 2 | 88.0 | 93.1 | 92.2 | **97.4** |
| **- Cos** | **88.3** | **93.4** | **92.3** | 97.3 |

*Table 1. Features' contributions on a 10-fold cross validation.*

---

[2] Ranking is calculated with Local Mutual Information (Evert, 2005)
[3] Entropy was calculated using the formula in Santus et al. (2014a)

[4] More info: https://github.com/esantus/. This work is partially supported by HK PhD Fellowship Scheme under PF12-13656.